\documentclass[10pt,twocolumn,letterpaper]{article}

\usepackage{cvpr}
\usepackage{times}
\usepackage{epsfig}
\usepackage{graphicx}
\usepackage{amsmath}
\usepackage{amssymb}
\usepackage{mathrsfs}
\usepackage[numbers]{natbib}
\usepackage[usenames,dvipsnames,svgnames,table]{xcolor}
\usepackage{booktabs}
\usepackage{soul}
\usepackage{xspace}
\usepackage{algorithm}
\usepackage[noend]{algpseudocode}
\usepackage{subcaption}
\usepackage[symbol]{footmisc}

\usepackage[pagebackref=true,breaklinks=true,letterpaper=true,colorlinks,bookmarks=false]{hyperref}

\graphicspath{{images/}}

\newcommand{\Eb}{Event-based\ }
\newcommand{\eb}{event-based\ }

\newcommand{\transpose}{^\intercal }
\newcommand{\bx}{\mathbf{x}}

\newcommand{\bz}{\mathbf{z}}

\newcommand{\bh}{\mathbf{h}}
\newcommand{\bH}{\mathbf{H}}

\newcommand{\M}{\mathcal{M}}
\newcommand{\N}{\mathcal{N}}
\newcommand{\E}{\mathcal{E}}
\newcommand{\T}{\mathcal{T}}
\newcommand{\C}{\mathcal{C}}
\newcommand{\R}{\mathbb{R}}

\newcommand{\hats}{{\it HATS\ }}
\newcommand{\hots}{{\it HOTS\ }}
\newcommand{\hfirst}{{\it H-First\ }}
\newcommand{\snn}{{\it SNN\ }}
\newcommand{\gsnn}{{\it Gabor-SNN\ }}
\newcommand{\nmnist}{N-MNIST\xspace}
\newcommand{\ncaltech}{N-Caltech101\xspace}
\newcommand{\ncars}{N-CARS\xspace}
\newcommand{\mnistdvs}{MNIST-DVS\xspace}
\newcommand{\cifardvs}{CIFAR10-DVS\xspace}

\cvprfinalcopy %

\def\cvprPaperID{1083} %

\ifcvprfinal\fi
\begin{document}

\title{HATS: Histograms of Averaged Time Surfaces for Robust \\Event-based Object Classification}

\author{Amos Sironi$^{1}$\thanks{This work was supported in part by the EU H2020 ULPEC project (grant agreement number 732642)}
	, Manuele Brambilla$^{1}$, Nicolas Bourdis$^{1}$, Xavier Lagorce$^{1}$, Ryad Benosman$^{1,2,3}$ \and
	$^{1}${\sc\Large prophesee}, Paris, France \quad
	$^{2}$Institut  de  la  Vision,  {\sc\Large upmc},  Paris, France \and
	$^{3}$University of Pittsburgh Medical Center / Carnegie Mellon University\and
	{\tt\small \{asironi, mbrambilla, nbourdis, xlagorce\}@prophesee.ai\ ryad.benosman@upmc.fr}
}

\maketitle
\thispagestyle{empty}

\begin{abstract}
	Event-based cameras have recently drawn the attention of the Computer Vision community thanks to their advantages 
	in terms of high temporal resolution, low power consumption and high dynamic range, compared to traditional frame-based cameras. 
	These properties make \eb cameras an ideal choice for autonomous vehicles, robot navigation or UAV vision, among others. 
	However, the accuracy of \eb object classification algorithms, 
	which is of crucial importance for any reliable system working in real-world conditions, 
	is still far behind their \textit{frame-based} counterparts.
	Two main reasons for this performance gap are: 
	1.\ The lack of effective low-level representations and architectures for \eb object classification and 
	2.\ The absence of large real-world \eb datasets.
	In this paper we address both problems. 
	First, we introduce a novel \eb feature representation together with a new machine learning architecture. %
	Compared to previous approaches, we use local memory units to efficiently leverage past temporal information 
	and build a robust \eb representation. 
	Second, we release the first large real-world \eb dataset for object classification.
	We compare our method to the state-of-the-art with extensive experiments, 
	showing better classification performance and real-time computation.
	
\end{abstract}

\section{Introduction}
\label{sec:introduction}

This paper focuses on the problem of object classification using the output 
of a neuromorphic asynchronous \eb camera \cite{Delbruck89,Delbruck10,Posch14}. 
\Eb cameras offer a novel path to Computer Vision 
by introducing a fundamentally new representation of visual scenes, 
with a drive towards real-time and low-power algorithms.

\begin{figure}
\centering
\begin{tabular}{@{\hspace{0.0cm}}c@{\hspace{0cm}}c@{\hspace{2pt}}}%
\multicolumn{2}{l}{\hspace{3mm}\includegraphics[width=0.85\linewidth]{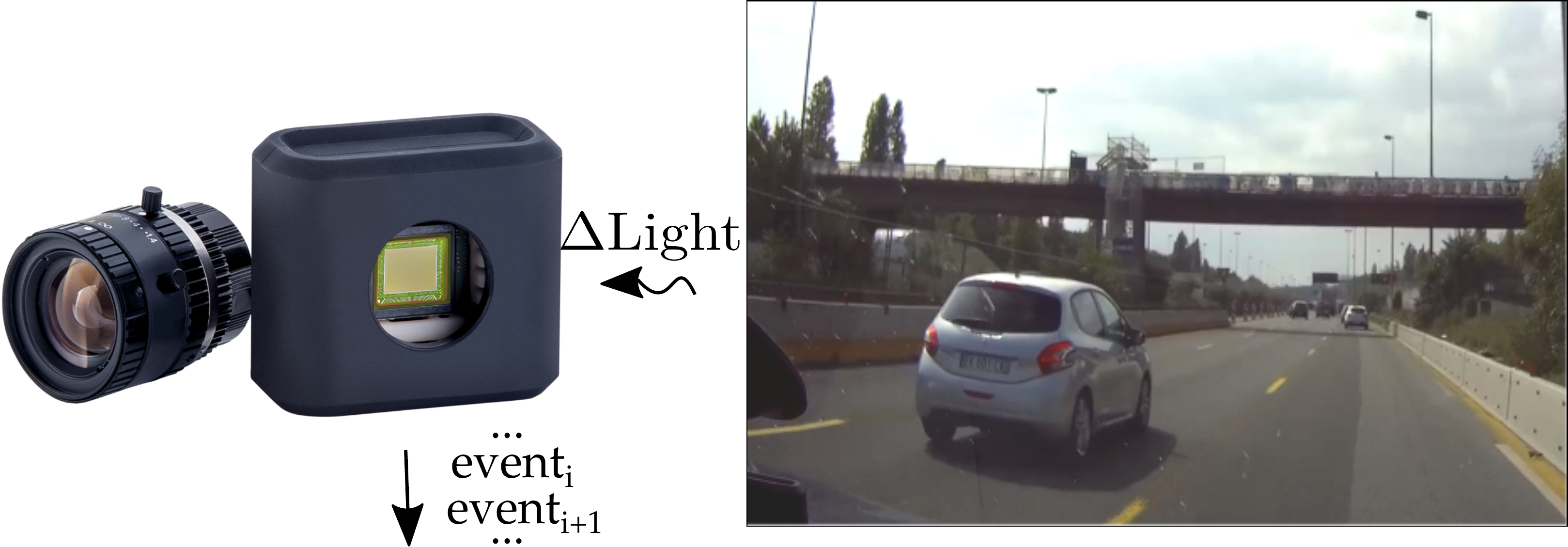}}%
\\
\multicolumn{2}{l}{\hspace{-1mm}\includegraphics[width=0.9\linewidth]{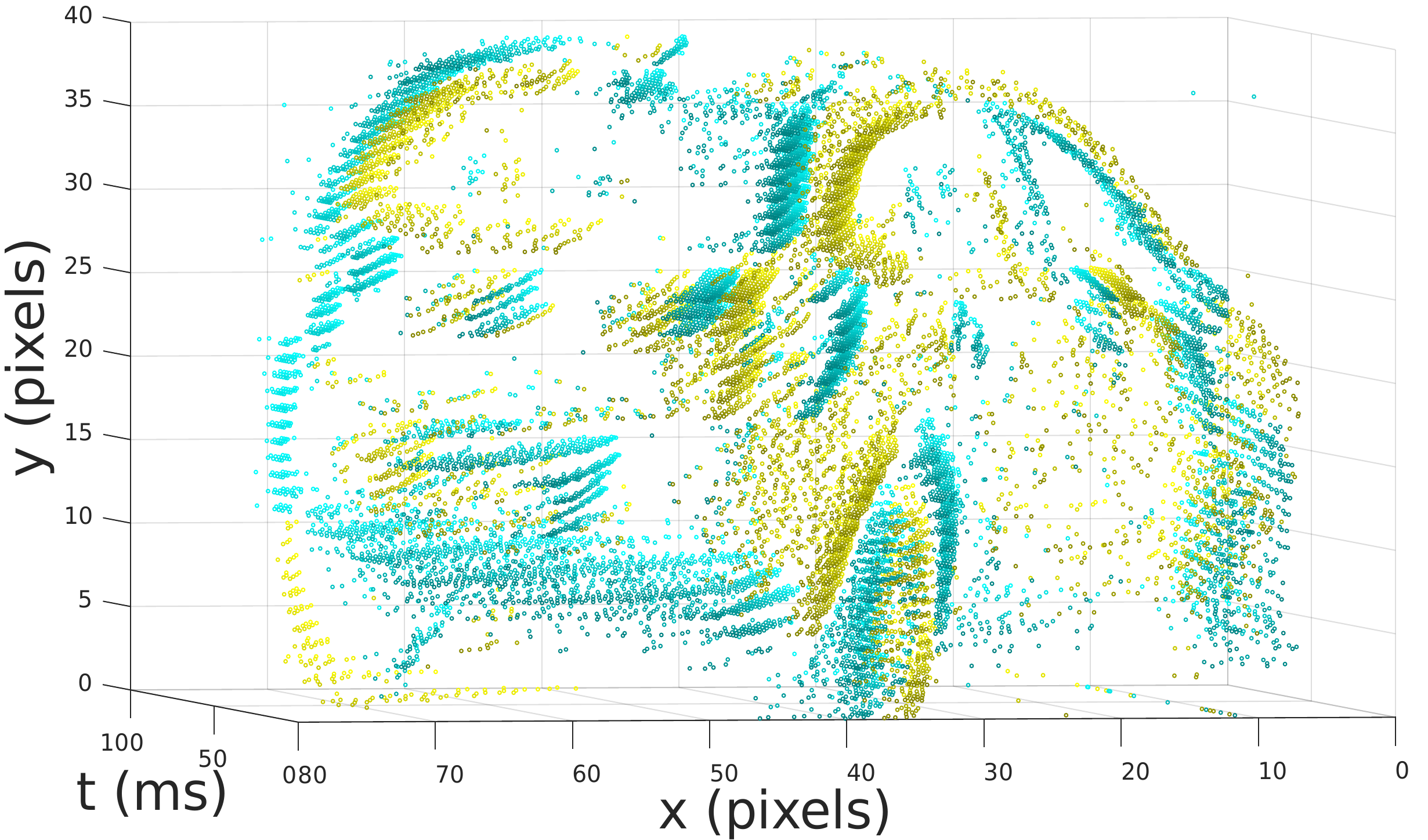}}%
\end{tabular}
\vspace{-2mm}
\caption{%
Pixels of an \eb camera asynchronously generate events as soon as a 
contrast change is detected in their field of view. As a consequence, the output of an \eb camera can be extremely sparse  
and with time resolution of order of microseconds. 
Because of the asynchronous nature of the data and the high resolution of the temporal component of the events, compared to the spatial one,
standard Computer Vision methods can not be directly applied.
 {\bf Top}: An \eb camera (left) recording a natural scene (right).
  {\bf Bottom}: Visualization of the events stream generated by a moving object. ON and OFF events (Sec.~\ref{sec:event-based_cameras}) are represented 
  by yellow and cyan dots respectively. %
  This figure, as most of the figures in this paper, is best seen in color.
}
\vspace{-5mm}
\label{fig:intro}
\end{figure}

Contrary to standard frame-based cameras, which rely on a pre-defined acquisition rate, in \eb cameras, 
individual pixels asynchronously emit \textit{events} when they observe a sufficient change of the local illuminance intensity (Figure~\ref{fig:intro}).

This new principle leads to significant reduction of memory usage and of power consumption 
and the information contained in standard videos of hundreds megabytes
can be naturally compressed in an event stream of few hundreds kilobytes~\cite{Lichtsteiner08,Posch11,Serrano13}.
Additionally, the time resolution of \eb cameras is orders of magnitude higher than frame-based cameras, 
reaching up to hundreds of microseconds. Finally, thanks to their logarithmic sensitivity to illumination changes, %
\eb cameras also have a much larger dynamic range, exceeding 120dB~\cite{Posch11}.
These characteristics make \eb cameras particularly interesting for applications with strong constraints
on latency (\eg autonomous navigation), power consumption (\eg UAV vision and IoT), or bandwidth (\eg tracking and surveillance).

However, due to the novelty of the field, the performance of \eb systems in real-world conditions 
is still inferior to their frame-based counterparts~\cite{Krizhevsky12,Shelhamer17}.
We argue that two main limiting factors of \eb algorithms are:
1. the limited amount of work on low-level feature representations and architectures for \eb object classification;
2. the lack of large \eb datasets acquired in real-world conditions.
In this work, we make important steps towards the solution of both problems.

We introduce a new \eb scalable machine learning architecture,
relying on a low-level operator called Local Memory Time Surface. 
A time surface is a spatio-temporal representation of activities around an 
event relying on the arrival time of events from neighboring pixels~\cite{Lagorce17}.
However, the direct use of this information is sensitive to noise and non-idealities of the sensors. 
By contrast, we emphasize the importance of using the information carried by past events 
to obtain a robust representation. 
Moreover, we show how to efficiently store and access this past information
by defining a new architecture based on local memory units,
where neighboring pixels share the same memory block. 
In this way, the Local Memory Time Surfaces can be efficiently combined 
into a higher-order representation, which we call Histograms of Averaged Time Surfaces.

This results in an \eb architecture which is significantly faster and more accurate than existing ones~\cite{Lagorce17,Lee16,Neil16}.
Driven by brain-like asynchronous event based computations, 
this new architecture offers the perspective of a new class of machine learning algorithms 
that focus the computational effort only on active parts of the network.

Finally, motivated by the importance of large-scale datasets for the recent progress of Computer Vision systems \cite{Deng09,Krizhevsky12,Lin14},
we also present a new real-world \eb  dataset dedicated to car classification.
This dataset is composed of about 24k samples acquired from a car driving in urban and motorway environments.
These samples were annotated using a semi-automatic protocol, which we describe below.
To the best of our knowledge this is the largest labeled \eb dataset acquired in real-world conditions.

We evaluate our method on our new \eb dataset and on four other challenging ones.
We show that our method reaches higher classification rates and faster computation times than existing \eb algorithms.

\section{Event-based camera}
\label{sec:event-based_cameras}

Conventional cameras encode the observed scene by producing dense information at a fixed frame-rate. 
As explained in Sec.~\ref{sec:introduction}, this is an inefficient way to encode natural scenes. 
Following this observation, a variety of \eb cameras~\cite{Lichtsteiner08,Posch11,Serrano13} 
have been designed over the past few years, with the goal to encode the observed scene adaptively, based on its content.

In this work, we consider the ATIS camera~\cite{Posch11}.
The ATIS camera contains an array of fully asynchronous pixels, each composed of an illuminance \textit{relative change detector} 
and a \textit{conditional exposure measurement block}.
The relative change detector reacts to changes in the observed scene, 
producing information in the form of \textit{asynchronous address events}~\cite{Boahen00}, 
known henceforth as \textit{events}. Whenever a pixel detects a change in illuminance intensity, 
it emits an event containing its x-y position in the pixel array, the microsecond timestamp of the observed change and its polarity:
i.e.\ whether the illuminance intensity was increasing (ON events) or decreasing (OFF events).
The conditional exposure measurement block measures the absolute luminous intensity observed by a pixel~\cite{Orchard14}. 
In the ATIS, the measurement itself is not triggered at fixed frame-rate, 
but only when a change in the observed scene is detected by the relative change detector. 

In this work, the luminous intensity measures from the ATIS camera were used only to generate ground-truth 
annotations for the dataset presented in Sec.~\ref{sec:datasets}.
By contrast, the object classification pipeline was designed to operate on change-events only, 
in order to support generic event-based cameras, whether or not they include the ATIS feature to generate grey levels.  
In this way, any event-based camera can be used to demonstrate the potential of our approach, 
while leaving the possibility for further improvement when gray level information is available~\cite{Liu16}.

\section{Related work}
\label{sec:related_work}

In this section, we first briefly review frame-based object classification,
then we describe previous work on \eb features and object classification. 
Finally, we discuss existing \eb datasets.
\paragraph{Frame-based Features and Object Classification}

There is a vast literature on spatial~\cite{Lowe99, Dalal05, Sivic09, Rublee11}
and spatio-temporal~\cite{Laptev05,Wang09,Scherer10} feature descriptors for frame-based Computer Vision. 
Early approaches mainly focus on hand-crafting feature representations for a given problem by using domain knowledge. 
Well-designed features combined with shallow classifiers have driven research in object recognition for many decades~\cite{Viola04,Dalal05,Dollar12}
and helped understanding and modeling important properties of the object classification problem, 
such as local geometric invariants, color and light properties, etc.~\cite{Witkin84,Arbelaez11}.  

In the last few years, the availability of large datasets~\cite{Deng09,Lin14} and effective learning algorithms~\cite{Lecun89,Kingma14,Srivastava14}  
shifted the research direction towards data driven learning of feature representations~\cite{Bengio13,Hinton06}. 
Typically this is done by optimizing the weights of several layers of elementary feature extraction operations,
such as spatial convolutions, pixel-wise transformations, pooling etc.
This allowed an impressive improvement in the performance of image classification approaches and many others Computer Vision problems~\cite{Krizhevsky12,Shelhamer17,Xie17}.
Deep Learning models, although less easily interpretable, also allowed understanding higher order geometrical properties of classical problems~\cite{Bruna13}.

By contrast, the work on \eb Computer Vision is still in its early stages and it is unclear which feature representations 
and architectures are best suited for this problem. 
Finding adequate low-level feature operations is a fundamental topic both for understanding the properties of \eb problems
and also for finding the best architectures and learning algorithms to solve them.

\paragraph{\Eb Features and Object Classification}
The majority of prior work on \eb features focused on detecting and tracking stable features adapted to Simultaneous Localization and Mapping 
applications~\cite{Kim16, Rebecq17}.
Corner detectors, have been defined in \cite{Clady15, Vasco16, Mueggler17}, while the works of~\cite{Seifozzakerini16,Brandli16} focused on edge and line extraction.

Recently,~\cite{Clady17} introduced a feature descriptor based on local distributions of optical flow and applied it to corner detection and gesture recognition.
It is inspired by its frame-based counterpart~\cite{Chaudhry09}, but in~\cite{Clady17} the algorithm for computing the optical flow 
relies on the temporal information carried by the events. 
One limitation of~\cite{Clady17} is that the quality of the descriptor strongly depends on the quality of the flow.
As a consequence, it loses accuracy in presence of noise or poorly contrasted edges.

\Eb classification algorithms can be divided in two categories: unsupervised learning methods and supervised ones.
Most unsupervised approaches train artificial neural networks 
by reproducing or imitating the learning rules observed in biological neural networks~\cite{Gutig06,Masquelier07,Linares11,Bichler12,Sheik13,Marti16}.
Supervised methods~\cite{Oconnor13,Lagorce15a,Li16,Peng17}, similar to what is done in frame-based Computer Vision, try to optimize the weights of artificial networks by minimizing a smooth error function.

The most commonly used architectures for \eb cameras are Spiking Neural Networks (SNN)~\cite{Bohte02,Russell10,Kasabov13,Diehl15,Cao15,Zhao15,Neil16}.
SNN are a promising research field; however, their performance is limited by the discrete nature of the events, 
which makes it difficult to properly train a SNN with gradient descent. 
To avoid this, some authors \cite{Orchard15c} use predefined Gabor filters as weights in the network.
Others propose to first train a conventional Convolutional Neural Networks (CNN) and then to convert the weights to a SNN~\cite{Cao15,Rueckauer16}.  
In both cases, the obtained solutions are suboptimal and typically the performance is lower than conventional CNNs on frames. 
Other methods consider a smoothed version of the transfer function of a SNN and directly optimize it~\cite{Lee16,Neil16b,Stromatias17}.
The convergence of the corresponding optimization problem is still very difficult to obtain and typically only few layers and small networks can be trained. 

Recently,~\cite{Lagorce17} proposed an interesting alternative to SNNs
by introducing a hierarchical representation based on the definition of \textit{Time Surface}.
In~\cite{Lagorce17}, learning is unsupervised and performed by clustering time surfaces at each layer,
while the last layer sends its output to a classifier. 
The main limitations of this method are its high latency, 
due to the increasing time window needed to compute the time surfaces 
and the high computational cost of the clustering algorithm.

We propose a much simpler yet effective feature representation. 
We generalize time surfaces by introducing a memory effect in the network by storing the information carried by past events. %
We then build our representation by applying a regularization scheme both in time and space 
to obtain a compact and fast representation. 
Although the architecture is scalable, we show that once a memory process is introduced, 
a single layer is sufficient to outperform a multilayer approach directly relying on time surfaces. 
This reduces computation, but more importantly, adds more generalization and robustness to the network.
\vspace{-4mm}

\paragraph{\Eb Datasets}

An issue of previous work on \eb object classification is that the proposed solutions are tested either on very small datasets~\cite{Serrano15,Lagorce17}, 
or on datasets generated by converting standard videos or images to an \eb representation~\cite{Orchard15,Hu16,Li17}.
In the first case, the small size of the test set prevents an accurate evaluation of the methods.
In the second case, the dataset size is large enough to create a valid tool for testing new algorithms.
However, since the datasets are generated from static images, the real dynamics of a scene and the temporal resolution 
of \eb cameras can not be fully employed
and there is no guarantee that a method tested on this kind of artificial data will behave similarly in real-world conditions.

The authors of~\cite{Mueggler17b} released an \eb dataset adapted to test visual odometry algorithms. 
Unfortunately, this dataset does not contain labeled information for an object recognition task.

The need of large real-world datasets is a major slowing factor for \eb vision \cite{Tan15}. 
By releasing a new labeled real-world event-based dataset, and defining an efficient 
semi-automated protocol based on a single event-based camera, 
we intend to accelerate progress toward a robust and accurate event-based object classifier.

\begin{figure*}[thpb]
	\centering
	\begin{tabular}{ccc}%
		\multicolumn{3}{c}{
		\includegraphics[width=0.95\linewidth]{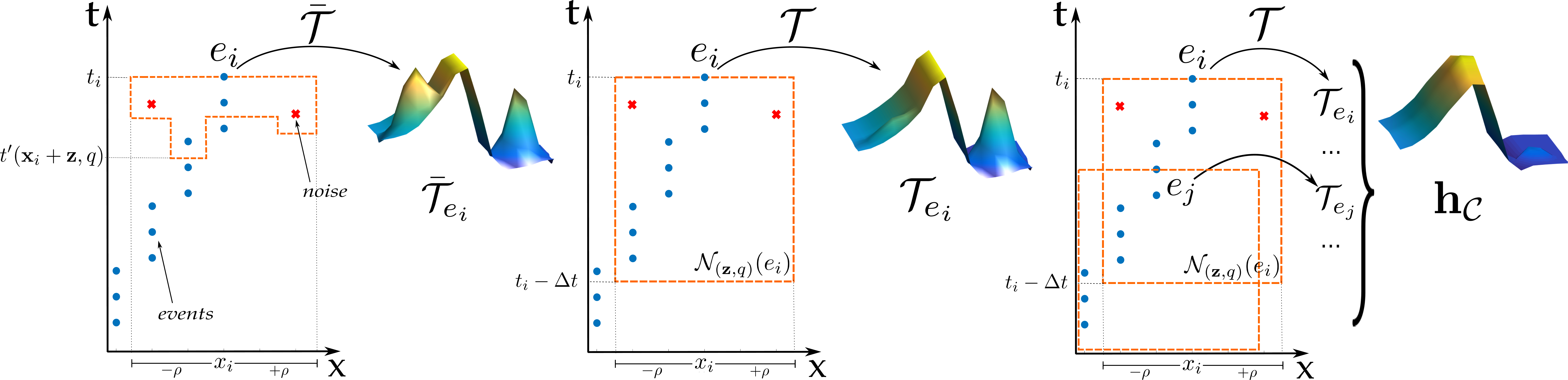}
		}\\
		{\hspace{2.15cm} \bf (a)} & {  \hspace{4.3cm}  {\bf (b)} } & {\hspace{0.55cm}  \bf (c)} 
	\end{tabular}
	\vspace{0.1em}
	\caption{Time surface computation around an event $e_i$, in presence of noise. 
		Noisy events are represented as red crosses, non-noisy events as blue dots. 
		For clarity of visualization only the $x$-$t$ component of the event stream and a single polarity are shown.
		{\bf (a)} In~\cite{Lagorce17} the time surface $\bar{\T}_{e_i}$ (Eq.~\eqref{eq:time_surface})
		is computed by considering only the times $t'(\bx_i+\bz,q)$ of the last
		events in a neighborhood of $e_i$ (orange dashed line). 
		As a consequence, noisy events can have a large weight in $\bar{\T}_{e_i}$. 
		This is visible from the spurious peaks in the surface $\bar{\T}_{e_i}$.
		{\bf (b)} By contrast, the definition of Local Memory Time Surface ${\T}_{e_i}$ of~Eq.~\eqref{eq:memory_time_surface}, 
		considers the contribution of all past events in a spatio-temporal window $\N_{(\bz,q)}{(e_i)}$. In this way, the ratio of noisy events 
		considered to compute $\T$ is smaller and the result better describes the real dynamics of the underling stream of events.
		{\bf (c)} The time surface can be further regularized by spatially averaging the time surfaces for all the events 
		in a neighborhood (Eq.~\eqref{eq:averaged_histogram}). Thanks to both the spatial and temporal regularization, 
		the contribution of noise is almost completely suppressed.
	}
	\vspace{-4mm}
	\label{fig:time_surface_example}
\end{figure*}

\section{Method}
\label{sec:method}

In this section, we formalize the \eb representation of visual scenes and describe our \eb architecture for object classification.

\subsection{Time Surfaces}\label{subsec:time_surface}
Given an \eb sensor with pixel grid size $M\times N$, a stream of events is given by a sequence 
\begin{equation}
	\E = \{e_i\}_{i=1}^{I},\ \text{with}\ e_i = (\bx_i, t_i,p_i),
	\label{eq:events_stream}
\end{equation}
where $\bx_i = (x_i,y_i)\in [1,\ldots,M]\times[1,\ldots,N]$ are the coordinates of the pixel generating the event, 
$t_i\geq0$ the timestamp at which the event was generated, with $t_i \leq t_j$ for $i<j$, and $p_i\in\{-1,1\}$ the \textit{polarity} of the event, 
with $-1,1$ meaning respectively OFF and ON events, and $I$ is the number of events.
From now on we will refer to individual events by $e_i$ and to a sequence of events by $\{e_i\}$. 

In \cite{Lagorce17}, the concept of time surface is introduced to describe local spatio-temporal patterns around an event. 
A time surface can be formalized as a local spatial operator acting on an event
$e_i$ by $\bar{\T}_{e_i}(\cdot,\cdot) : [-\rho,\rho]^2\times\{-1,1\} \rightarrow \R$, 
where $\rho$ is the radius of the spatial neighborhood used to compute the time surface. 

For an event $e_i = (\bx_i, t_i, p_i)$, and $(\bz,q)\in[-\rho,\rho]^2\times\{-1,1\}$, $\bar{\T}_{e_i}$ is given by
\begin{equation}
	\bar{\T}_{e_i}(\bz,q) = \left\{
	\begin{array}{ll}
		e^{-\frac{t_i - t'(\bx_i+\bz,q)}{\tau}} & \quad \text{if } p_i = q \\
		0                                       & \quad \text{otherwise.}  
	\end{array} \right.
	\label{eq:time_surface}
\end{equation}

Where $t'(\bx_i+\bz,q)$ is the time of the last event with polarity $q$ received from pixel $\bx_i+\bz$ (Fig.~\ref{fig:time_surface_example}(a)),
and $\tau$ is a decay factor giving less weight to events further in the past.
Intuitively, a time surface encodes the dynamic context in a neighborhood of an event,
hence providing both temporal and spatial information.
Therefore, this compact representation of the content of the scene can be useful to classify different patterns.

\begin{figure*}[tphb]
\centering
  \includegraphics[width=0.95\linewidth,height=0.25\linewidth]{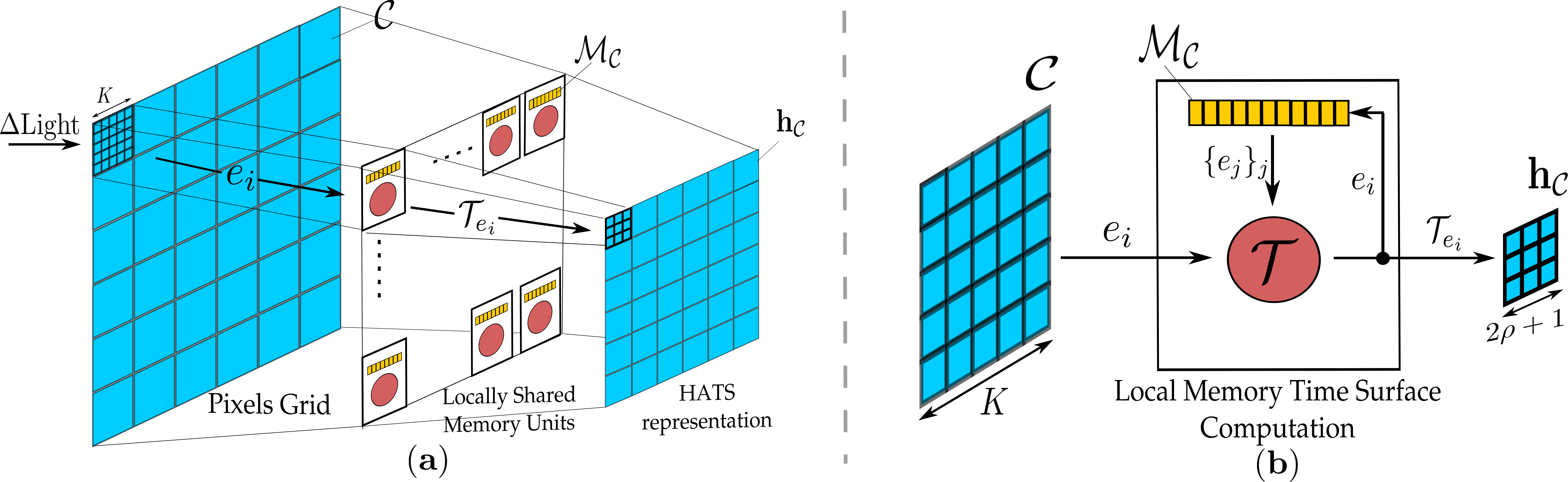}
\vspace{0.1em}
\caption{Overview of the proposed architecture. {\bf (a)} The pixel grid is divided into cells $\C$ of size $K\times K$. When a change of light is detected by a pixel, 
an event $e_i$ is generated. Then, the time surface $\T_{e_i}$ is computed and used to update the histogram $\bh_{\C}$. The \hats representation is obtained by the concatenation 
of the histograms $\bh_{\C}$. {\bf (b)} Detail of the Local Memory Time Surface computation using local memory units. For each input event, 
the time surface of Eq.~\eqref{eq:memory_time_surface} is computed by using the past events $e_j$'s stored in the cell's local memory unit $\M_C$ (Sec.~\ref{subsec:implementation}).
After computation, $\T_{e_i}$ is used to update the histogram $\bh_{\C}$ of the corresponding cell, while event $e_i$ is added to the memory unit.
For simplicity, the polarity of the event and the normalization of the histograms are not considered in the scheme.
}
\label{fig:method_overview}
\end{figure*}

\subsection{Local Memory Time Surfaces}\label{subsec:memory_time_surface}
To build the feature representation, we start by generalizing the time surface $\bar{\T}_{e_i}$ of Eq.~\eqref{eq:time_surface}.
As shown in Fig.~\ref{fig:time_surface_example}(a) using only the time $t'(\bx_i+\bz,q)$ of the last event received in the neighborhood of the time surface pixel $\bx_i$,
leads to a descriptor which is too sensitive to noise or small variations in the event stream.  

To avoid this problem, we compute the time surface by considering the history of the events in a temporal window of size $\Delta t$.
More precisely, we define a local memory time surface $\T_{e_i}$ as 
\begin{equation}
	\T_{e_i}(\bz,q) = \left\{
	\begin{array}{ll}
		\sum_{e_j\in\N_{(\bz,q)}(e_i)}{e^{-\frac{t_i - t_j}{\tau}} } & \ \text{if } p_i = q \\
		0                                                            & \ \text{otherwise,}  
	\end{array} \right.
	\label{eq:memory_time_surface}
\end{equation}
where
\begin{equation}
	\small{
		\N_{(\bz,q)}(e_i)=\{e_j: \bx_j=\bx_i+\bz, t_j \in [t_i-\Delta t, t_i), p_j = q\}.
		}
		\label{eq:spatio_temporal_neighbor}
	\end{equation}
	As shown in Fig.~\ref{fig:time_surface_example}(b), 
	this formulation more robustly describes the real dynamics of the scene while resisting noise and small variations of events.
	In the supplementary material we compare the results obtained by using Eq.~\eqref{eq:time_surface} or Eq.~\eqref{eq:memory_time_surface} 
	on an object classification task, showing the advantage of using the local memory formulation to achieve better accuracy.
					
	The name \textit{Local Memory Time Surfaces} comes from the fact that past events $\{e_j\}$ in $\N_{(\bz,q)}(e_i)$
	need to be stored in memory units in order to prevent the algorithm from `forgetting' past information.
	In Sec.~\ref{subsec:implementation}, we will describe how memory units can be shared efficiently by neighboring pixels.
	In this way, we can compute a robust feature representation without significant increase in memory requirements.

	\subsection{Histograms of Averaged Time Surfaces}\label{subsec:hats}
	The local memory time surfaces of Eq.~\eqref{eq:memory_time_surface} is the elementary spatio-temporal operator we use in our approach.
	In this section, we describe how this new type of time surface can be used to define 
	a compact representation of an event stream useful for object classification.

	Inspired by~\cite{Dalal05} in frame-based vision, we group adjacent pixels in cells $\{\C_l\}_{l=1}^{L}$ of size $K\times K$.
	Then, for each cell $\C$, we sum the components of the time surfaces computed on events from $\C$ into histograms. 
	More precisely, for a cell $\C$ we have:
					
	\begin{equation}
		\bar{\bh}_{\C}(\bz,p) = \sum_{e_i\in \C}{\T_{e_i}(\bz,p)}, %
		\label{eq:cell_histogram}
	\end{equation}
	where, with an abuse of notation, we write $e_i\in \C$ if and only if pixel coordinates 
	$(x_i,y_i)$ of the event belong to $\C$.
					
	A characteristic of \eb sensors is that the amount of events generated by a moving object is proportional to its contrast: 
	higher contrast objects generate more events than low contrast objects. 
	To make the cell descriptor more invariant to contrast, we therefore normalize $\bar{\bh}$ by the number of events $|\C|$ contained in the spatio-temporal window used to compute it. This results in the averaged histogram:
	\begin{equation}
		\bh_{\C}(\bz,p) = \frac{1}{|\C|}\bar{\bh}_{\C}(\bz,p) = \frac{1}{|\C|}\sum_{e_i\in \C}{\T_{e_i}(\bz,p)}.
		\label{eq:averaged_histogram}
	\end{equation}
	An example of a cell histogram $\bh_{\C}(\bz,p)$ is shown in Fig.~\ref{fig:time_surface_example}(c).
	Given a stream of events, our final descriptor, which we call \hats for Histograms of Averaged Time Surfaces, is given by concatenating every $\bh_{\C}$, for all positions $\bz$, polarities and cells $1,\ldots,L$:
	\begin{equation}
		\bH(\{e_i\}) = [\bh_{\C_1},\ldots,\bh_{\C_L}]\transpose.
		\label{eq:hats}
	\end{equation}
	Fig.~\ref{fig:method_overview}(a) shows an overview of our method.

	Similarly to standard Computer Vision methods, we can further group adjacent cells into blocks and perform a block-normalization scheme to obtain more invariance to velocity and contrast~\cite{Dalal05}. 
	In Sec.~\ref{sec:experiments}, we show how this simple representation obtains higher accuracy for \eb object classification compared to previous approaches.
					
	\begin{algorithm}[t]
		\caption{\hats with shared memory units}
		\label{alg:hats}
		\begin{algorithmic}[1]
			\State Input: Events $\E = \{e_i\}_{i=1}^{I}$
			Parameters: $\rho,\Delta t, \tau, K$
			\State Output: \hats representation $\bH(\{e_i\})$
			\State Initialize: $\bh_{\C_l} =\textbf{0},\ |\C_l| =0,\ \M_{\C_l} = \emptyset, \text{for all}\ l$   
			\For{$i = 1,\ldots,I$}
			\State $\C_l \leftarrow \text{getCell}(x_i,y_i)$
			\State $\T_{e_i} \leftarrow \text{computeTimeSurface}(e_i,\M_{\C_l})$ %
			\State $\bh_{\C_l} \leftarrow \bh_{\C_l} + \T_{e_i}$
			\State $\M_{\C_l} \leftarrow \M_{\C_l} \cup e_i$
			\State $|\C_l| \leftarrow |\C_l| + 1$
			\EndFor
			\State return $\bH = [\bh_{\C_1}/|\C_1|,\ldots,\bh_{\C_L}/|\C_L|]\transpose $
		\end{algorithmic}
	\end{algorithm}
	\subsection{Architecture with Locally Shared Memory Units}\label{subsec:implementation}

	Irregular access in \eb cameras is a well known limiting factor for designing efficient event-based algorithms. 
	One of the main problems is that the use of standard hardware accelerations, 
	such as GPU, is not trivial due to the sparse and asynchronous nature of the events. 
	For example, accessing spatial neighbors on contiguous memory blocks can impose significant overheads when processing event-based data.

	The architecture computing the \hats representation allows to overcome this memory access issue (Fig.~\ref{fig:method_overview}).
	From Eq.~\eqref{eq:cell_histogram} we notice that for every incoming event $e_i$, 
	we need to iterate over all events in a past spatio-temporal neighborhood.
	Since, for small values of $\rho$, most of the past events would not be in the neighborhood of $e_i$,
	looping through the entire temporally ordered event stream would be prohibitively expensive and inefficient.
	To avoid this, we notice that, for $\rho \approx K$, 
	the events falling in the same cell $\C$, will share most of the neighbors $\N_{(\bz,q)}$ used to compute Eq.~\eqref{eq:memory_time_surface}.
	Following this observation, for every cell, we define a shared memory unit $\M_C$, where past events relevant for $\C$ are stored.
	In this way, when a new event arrives in $\C$, we update Eq.~\eqref{eq:cell_histogram} by only looping through $\M_C$,
	which contains only the relevant past events to compute the Local Memory Time Surface of Eq.~\eqref{eq:memory_time_surface} (Fig.~\ref{fig:method_overview}(b)).
					
	Algorithm~\ref{alg:hats} describes the computation of \hats with memory units.
	Although this was not the scope of this paper, we notice that Algorithm~\ref{alg:hats} can be easily parallelized 
	and implemented in dedicated neuromorphic chips~\cite{Serrano06}.

\begin{figure*}[t]
  \centering
  \includegraphics[width=0.95\linewidth]{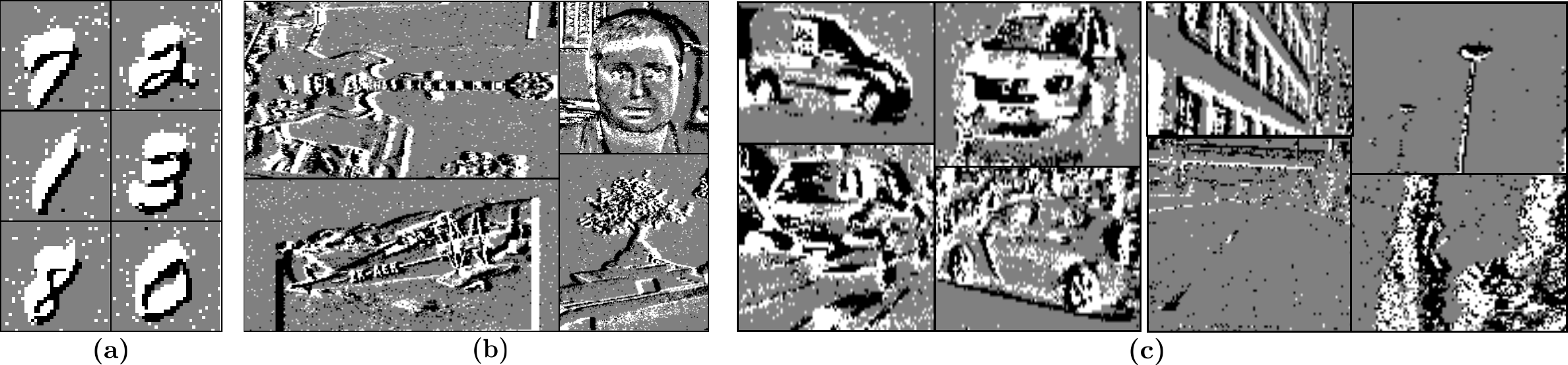}
\caption{Sample snapshots from the datasets used for the experiments of Sec.~\ref{sec:experiments}. 
The snapshots are obtained by cumulating 100ms of events. Black pixels represents OFF events, white pixels ON events.
 {\bf (a)} \nmnist Dataset. {\bf (b)} \ncaltech dataset. {\bf (c)} \ncars Dataset. Left: positive samples; Right: negative samples.
 Notice that the \nmnist and \ncaltech datasets have been generated by moving an \eb camera in front of a LCD screen displaying static images. 
 By contrast, our dataset has been acquired in real-world conditions, therefore it fully exploits the temporal resolution of the camera
 by capturing the real dynamics of the objects.}
\label{fig:datasets_examples}
\vspace{-5mm}
\end{figure*}

\section{Datasets}
\label{sec:datasets}

We validated our approach on five different datasets: four datasets generated by converting 
standard frame-based datasets to events (namely, the \nmnist~\cite{Orchard15}, \ncaltech~\cite{Orchard15},
\mnistdvs~\cite{Serrano13} and \cifardvs~\cite{Li17} datasets)
and a novel dataset, recorded from real-world scenes and introduced for the first time in this paper, which we call \ncars. 
We made the \ncars dataset publicly available for download at \texttt{http://www.prophesee.ai/dataset-n-cars/}.

\subsection{Datasets Converted from Frames}%

\nmnist, \ncaltech, \mnistdvs and \cifardvs are four publicly available datasets created by converting the popular 
frame-based MNIST~\citep{Lecun89}, Caltech101~\citep{Fei06} and CIFAR10~\cite{Krizhevsky09} to an \eb representation.

\nmnist and \ncaltech were obtained by displaying each sample image on an LCD monitor,
while an ATIS sensor (Section~\ref{sec:event-based_cameras}) was moving in front of it~\cite{Orchard15}.
Similarly, the \mnistdvs and \cifardvs datasets were created by displaying a moving image on a monitor and
recorded with a fixed DVS sensor~\cite{Serrano13}. 

In both cases, the result is a conversion of the images of the original datasets into a stream of events suited 
for evaluating event-based object classification. 
Fig.~\ref{fig:datasets_examples}(a,b) shows some representative examples of the datasets generated from frames, for the \nmnist and \ncaltech.

\subsection{Dataset Acquired Directly as Events: N-CARS}
\label{sec:sub_ncars}

The datasets described in the previous section are good datasets for a first evaluation of event-based classifiers.
However, since they were generated by displaying images on a monitor, 
they are not very representative of data from real-world situations. 
The main shortcoming results from the limited and predefined motion of the objects.

To overcome these limitations, we created a new dataset by directly recording objects in urban environments with an \eb sensor. 
The dataset was obtained with the following semi-automatic protocol. First, we captured approximately 80 minutes of video using 
an ATIS camera (Section \ref{sec:event-based_cameras}) mounted behind the windshield of a car. 
The driving was conducted in a natural way, without particular regards for video quality or content.
In a second stage, we converted gray-scale measurements from the ATIS sensor to conventional gray-scale images. 
We then processed them with a state-of-the-art object detector~\citep{Redmon16, ren15}, 
to automatically extract bounding boxes around cars and background samples.
Finally, the data was manually cleaned to ensure that the samples were correctly labeled. 
 
Since the gray-scale measurements have the same time resolution of the change detection events,
the gray-level images can be easily synchronized with the change detection events. 
Thus, the positions and timestamps of the bounding boxes can be directly used to extract the corresponding 
event-based samples from the full event stream. 
Thanks to our semi-automated protocol, we generated a two-class dataset composed of 12,336 car samples
and 11,693 non-cars samples (background). 
The dataset was split in 7940 car and 7482 background training samples,
and 4396 car and 4211 background testing samples. 
Each example lasts 100 milliseconds. More details on the dataset can be found in the supplementary material.

We called this new dataset \ncars. As shown in Fig.~\ref{fig:datasets_examples}(c) the \ncars is a challenging dataset,
containing cars at different poses, speeds and occlusions, as well as a large variety of background scenarios.

\section{Experiments}
\label{sec:experiments}

\subsection{\Eb Object Classification}
Once the features have been extracted from the events sequences of the database, the problem reduces to a conventional classification problem. 
To highlight the contribution of our feature representation to classification accuracy, we used a simple linear SVM classifier
in all our experiments. A more complex classifier, such as non-linear SVM or Convolutional Neural Networks, could be used to further improve the results.

The parameters for all methods were optimized by splitting the training set and using 20\%
of the data for validation. Once the best settings were found, the classifier was retrained on the whole training set.

We noticed little influence of the $\rho \text{ and } \tau$ parameters to accuracy, 
while small $K$'s improved performance for low resolution inputs.
When the input duration is larger than the value of $\Delta t$ used to compute the time
surfaces (Eq.~\ref{eq:spatio_temporal_neighbor}), we compute the features every 
$\Delta t$ and then stack them together.

The baselines methods we consider are \hots~\cite{Lagorce17}, \hfirst~\cite{Orchard15c} and Spiking Neural Networks ({\it SNN})~\cite{Lee16,Neil16}. 
For \hfirst we used the code provided by the authors online. %
For \snn we report the results previously published, when available, while for \hots we used our implementation of the method described in~\cite{Lagorce17}.
As with \hats features, we used a linear SVM on the features extracted with {\it HOTS}. 
Notice that this is in favour of {\it HOTS}, since linear SVM is a more powerful classifier than the one 
used by the authors~\cite{Lagorce17}.

Given that no code is available for SNN, we also compared our results with those of a 2-layer SNN architecture 
we implemented using predefined Gabor filters~\cite{Bovik90}. 
We then again train a linear SVM on the output of the network. We call this approach {\it Gabor-SNN}.
This allowed us to obtain the results for SNN when not readily available in the literature.

\vspace{-2mm}
\paragraph{Results on the Datasets Converted from Frames}%

\begin{table*}[htpb]
	\caption{Comparison of classification accuracy on datasets converted from frames. 
		Our method has the highest classification rate ever
		reported for an \eb classification method.}
	\vspace{-4mm}
	\begin{center}
		\tabcolsep=0.11cm
		\begin{tabular}{@{}l cccc@{}}
			\toprule
			\textbf{}                   & \textbf{{\nmnist}} & \textbf{{\ncaltech}} & \textbf{{\mnistdvs}} & \textbf{{\cifardvs}} \\
			\hline
			{\hfirst~\cite{Orchard15c}} & 0.712              & 0.054                & 0.595                & 0.077                \\
			{\hots~\cite{Lagorce17}}    & 0.808              & 0.210                & 0.803                & 0.271                \\
			\gsnn                       & 0.837              & 0.196                & 0.824                & 0.245                \\
			{\hats~(this work)}         & \textbf{0.991}     & \textbf{0.642}       & \textbf{0.984}       & \textbf{0.524}       \\
			\hline
			{Phased LSTM~\cite{Neil16}} & 0.973              & -                    & -                    & -                    \\
			{Deep SNN~\cite{Lee16}}     & 0.987              & -                    & -                    & -                    \\
			\bottomrule
		\end{tabular}%
	\end{center}
	\label{tab:nmnist_ncaltech}
	\vspace{-5mm}
\end{table*}

The results for the \nmnist, \ncaltech, \mnistdvs and \cifardvs datasets are given in Tab.~\ref{tab:nmnist_ncaltech}.
As it is usually done, we report the results in terms of classification accuracy. %
The complete set of parameters used for the methods are reported in the supplementary material.
 
Our method has the highest classification rate ever
reported for an \eb classification method. 
The performance improvement is higher for the more challenging \ncaltech and \cifardvs datasets.
\hots and a predefined \gsnn have similar performance, while the \hfirst 
learning mechanism is too simple to reach good performance.%

\vspace{-2mm}
\paragraph{Results on the \ncars Datasets}
For the \ncars dataset, the \hats parameters used are $K=10$, $\rho=3$ and $\tau=10^{9}\mu s$. 
In this case, block normalization was not applied because it did not improve results.
Since the \ncars dataset contains only two classes, cars and non-cars, we can consider it as a binary classification problem. 
Therefore, we also analyze the performance of the methods using ROC curves analysis~\cite{Fawcett06}.
The Area Under the Curve (AUC) and the accuracy (Acc.) for our method and the baselines are shown in Tab.~\ref{tab:ncars}, 
while the ROC curves are presented in the supplementary material.
\begin{table}[htpb]
	\caption{Comparison of classification results on the \ncars dataset. The table reports the global classification accuracy (Acc.) 
		and the AUC score (the higher the better). 
	Our method outperforms the baselines by a large margin.}
	\begin{center}
		\vspace{-4mm}
		\begin{tabular}{@{}l cc@{}}
			\toprule
			\textbf{\ncars}             & Acc.           & AUC            \\
			\hline
			{\hfirst~\cite{Orchard15c}} & 0.561          & 0.408          \\
			{\hots~\cite{Lagorce17}}    & 0.624          & 0.568          \\
			\gsnn                       & 0.789          & 0.735          \\
			{\hats~(this work)}         & \textbf{0.902} & \textbf{0.945} \\
			\bottomrule
		\end{tabular}%
	\end{center}
	\label{tab:ncars}
	\vspace{-4mm}
\end{table}

From the results, we see that our method outperforms the baselines by a large margin. 
The variability contained in a real-world dataset, such as the \ncars one, is too large for both the \hfirst and \hots
learning algorithms to converge to a good feature representation. A predefined \gsnn architecture has better accuracy than \hfirst and {\it HOTS}, 
but still 11\% lower than our method. 
The spatio-temporal regularization implemented in our method
is more robust to the noise and variability contained in the dataset.

\subsection{Latency and Computational Time}

Latency is a crucial characteristic for many applications requiring fast reaction time. 
In this section, we compare \hats, \hots and \gsnn in terms of their computational time and latency on the \ncars dataset. 
All methods are implemented in C++ and run on a laptop equipped with an Intel i7 CPU (64bits, 2.7GHz) and 16GB of RAM.

Tab.~\ref{tab:computational_times} compares the average computational times to process a sample. 
Average computational time per sample was computed by dividing the total time spent to compute the features on the full training set
by the number of training samples.
As we can see, our method is more than 20x faster than \hots and almost 40x times faster than a 2-layer SNN.
In particular our method is 13 times faster than real time. 
We also report the average number of events processed per second in Kilo-events per second (Kev/s).

Latency represents the time period used to accumulate evidence in order to reach a decision on the object class. 
In our case, this time period is given by the time window used to compute the features, as longer time windows results in higher latency.
Notice that with this definition, the latency is independent from both the computational time and the classification accuracy.

\begin{figure}
    \centering
    \begin{tabular}{@{\hspace{0cm}}c@{\hspace{0cm}}}%
      \includegraphics[width=0.95\linewidth]{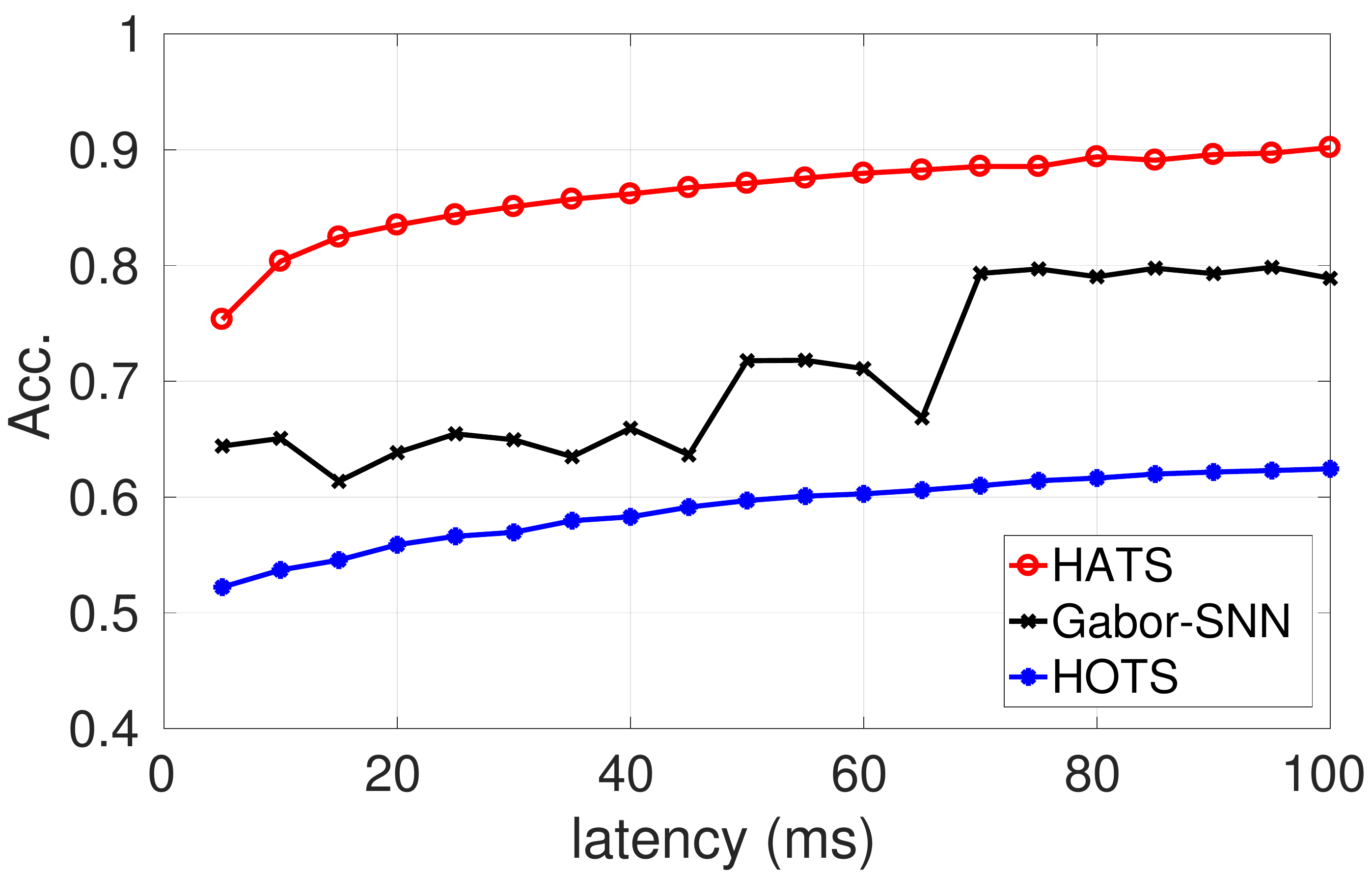}
    \end{tabular}
    \vspace{-2mm}
    \caption{Accuracy as a function of latency on the \ncars dataset.
    Our method is consistently more accurate than the baselines and already reaches better performance 
    by using only events contained in the first 10ms of the samples.}
    \vspace{-3mm}
    \label{fig:accuracy_vs_latency}
\end{figure}

There is a trade-off between latency and classification accuracy: on one side longer time periods yield more information at the cost of higher latency, 
on the other side they lead to risk of mixing dynamics from separate objects or even different dynamics from the same object.
We study this trade-off by plotting the accuracy as a function of the latency for the different methods (Fig.~\ref{fig:accuracy_vs_latency}). 
The results were averaged over 5 repetitions.
By using only 10ms of events, \hats has higher performance than the baselines applied to the full 100ms events stream. 
The performance of \hats does not completely saturate, probably due to the presence of cars with really small apparent motion in the dataset.

We also notice that the performance of \gsnn is unstable, especially for low latency. 
This is due to the spiking architecture of \gsnn for which small variations in the input of a layer can cause large differences at its output.

\begin{table}[htpb]
	\caption{Average computational times per sample (the lower the better) and average number of events processed per second, in Kilo-events per second Kev/s (the higher the better), on the \ncars dataset.
		Since each sample is 100ms long, our method is more than 13 times faster than real time, while \hots and \gsnn are respectively 1,5 and 2,8
	times slower than real time.}
	\vspace{-4mm}
	\begin{center}
		\tabcolsep=0.11cm
		\begin{tabular}{@{}l c@{} c@{}}
			\toprule
			\textbf{\ncars}          & Average Comp.        & Kev/s           \\
			\textbf{}                & Time per Sample (ms) &                 \\
			\hline
			{\hots~\cite{Lagorce17}} & 157.57               & 25.68           \\ %
			\gsnn                    & 285.95               & 14.15           \\
			{\hats~(this work)}      & \textbf{7.28}        & \textbf{555.74} \\ %
			\bottomrule
		\end{tabular}%
	\end{center}
	\label{tab:computational_times}
\end{table}
\vspace{-2mm}

\section{Conclusion and Future Work}
\label{sec:conclusion}
In this work, we presented a new feature representation for \eb object recognition 
by introducing the notion of Histograms of Averaged Time Surfaces.
It validates the idea that information is contained in the relative time between events, 
provided a regularization scheme is introduced to limit the effect of noise.
The proposed architecture makes efficient use of past information by using local memory units shared by neighboring pixels,
outperforming existing spike based methods in both accuracy and efficiency.

In the future, we plan to extend our method by using a feature representation also for the memory units, instead of using raw events. 
This could be done for example by training a network to learn linear weights to apply to the incoming time surfaces.

{\small
	\bibliographystyle{ieee}
	\bibliography{short,biblio}

\begin{thebibliography}{10}\itemsep=-1pt

\bibitem{Arbelaez11}
P.~Arbelaez, M.~Maire, C.~Fowlkes, and J.~Malik.
\newblock Contour detection and hierarchical image segmentation.
\newblock {\em TPAMI}, 2011.

\bibitem{Bengio13}
Y.~Bengio, A.~Courville, and P.~Vincent.
\newblock Representation learning: A review and new perspectives.
\newblock {\em TPAMI}, 2013.

\bibitem{Bichler12}
O.~Bichler, D.~Querlioz, S.~J. Thorpe, J.-P. Bourgoin, and C.~Gamrat.
\newblock Extraction of temporally correlated features from dynamic vision
  sensors with spike-timing-dependent plasticity.
\newblock {\em Neural Networks}, 2012.

\bibitem{Boahen00}
K.~A. Boahen.
\newblock Point-to-point connectivity between neuromorphic chips using
  address-events.
\newblock {\em IEEE Trans. Circuits Syst. II}, 2000.

\bibitem{Bohte02}
S.~M. Bohte, J.~N. Kok, and H.~La~Poutre.
\newblock Error-backpropagation in temporally encoded networks of spiking
  neurons.
\newblock {\em Neurocomputing}, 2002.

\bibitem{Bovik90}
A.~C. Bovik, M.~Clark, and W.~S. Geisler.
\newblock Multichannel texture analysis using localized spatial filters.
\newblock {\em TPAMI}, 1990.

\bibitem{Brandli16}
C.~Br{\"a}ndli, J.~Strubel, S.~Keller, D.~Scaramuzza, and T.~Delbruck.
\newblock Elised—an event-based line segment detector.
\newblock In {\em Event-based Control, Communication, and Signal Processing
  (EBCCSP), International Conference on}, 2016.

\bibitem{Bruna13}
J.~Bruna and S.~Mallat.
\newblock Invariant scattering convolution networks.
\newblock {\em TPAMI}, 2013.

\bibitem{Cao15}
Y.~Cao, Y.~Chen, and D.~Khosla.
\newblock Spiking deep convolutional neural networks for energy-efficient
  object recognition.
\newblock {\em IJCV}, 2015.

\bibitem{Chaudhry09}
R.~Chaudhry, A.~Ravichandran, G.~Hager, and R.~Vidal.
\newblock Histograms of oriented optical flow and binet-cauchy kernels on
  nonlinear dynamical systems for the recognition of human actions.
\newblock In {\em CVPR}, 2009.

\bibitem{Clady15}
X.~Clady, S.-H. Ieng, and R.~Benosman.
\newblock Asynchronous event-based corner detection and matching.
\newblock {\em Neural Networks}, 2015.

\bibitem{Clady17}
X.~Clady, J.-M. Maro, S.~Barr{\'e}, and R.~B. Benosman.
\newblock A motion-based feature for event-based pattern recognition.
\newblock {\em Frontiers in neuroscience}, 2017.

\bibitem{Dalal05}
N.~Dalal and B.~Triggs.
\newblock {Histograms of Oriented Gradients for Human Detection}.
\newblock In {\em CVPR}, 2005.

\bibitem{Delbruck10}
T.~Delbr{\"u}ck, B.~Linares-Barranco, E.~Culurciello, and C.~Posch.
\newblock Activity-driven, event-based vision sensors.
\newblock In {\em Proc. IEEE International Symposium on Circuits and Systems},
  2010.

\bibitem{Delbruck89}
T.~Delbr{\"u}ck and C.~Mead.
\newblock An electronic photoreceptor sensitive to small changes in intensity.
\newblock In {\em NIPS}, 1989.

\bibitem{Deng09}
J.~Deng, W.~Dong, R.~Socher, L.-J. Li, K.~Li, and L.~Fei-Fei.
\newblock {Imagenet: A Large-Scale Hierarchical Image Database}.
\newblock In {\em CVPR}, 2009.

\bibitem{Diehl15}
P.~U. Diehl, D.~Neil, J.~Binas, M.~Cook, S.-C. Liu, and M.~Pfeiffer.
\newblock Fast-classifying, high-accuracy spiking deep networks through weight
  and threshold balancing.
\newblock In {\em International Joint Conference on Neural Networks}, 2015.

\bibitem{Dollar12}
P.~Dollar, C.~Wojek, B.~Schiele, and P.~Perona.
\newblock Pedestrian detection: An evaluation of the state of the art.
\newblock {\em TPAMI}, 2012.

\bibitem{Fawcett06}
T.~Fawcett.
\newblock An introduction to roc analysis.
\newblock {\em Pattern recognition letters}, 2006.

\bibitem{Fei06}
L.~Fei-Fei, R.~Fergus, and P.~Perona.
\newblock One-shot learning of object categories.
\newblock {\em TPAMI}, 2006.

\bibitem{Gutig06}
R.~G{\"u}tig and H.~Sompolinsky.
\newblock The tempotron: a neuron that learns spike timing--based decisions.
\newblock {\em Nature neuroscience}, 2006.

\bibitem{Hinton06}
G.~E. Hinton and R.~R. Salakhutdinov.
\newblock Reducing the dimensionality of data with neural networks.
\newblock {\em Science}, 2006.

\bibitem{Hu16}
Y.~Hu, H.~Liu, M.~Pfeiffer, and T.~Delbruck.
\newblock Dvs benchmark datasets for object tracking, action recognition, and
  object recognition.
\newblock {\em Frontiers in neuroscience}, 2016.

\bibitem{Kasabov13}
N.~Kasabov, K.~Dhoble, N.~Nuntalid, and G.~Indiveri.
\newblock Dynamic evolving spiking neural networks for on-line spatio-and
  spectro-temporal pattern recognition.
\newblock {\em Neural Networks}, 2013.

\bibitem{Kim16}
H.~Kim, S.~Leutenegger, and A.~J. Davison.
\newblock Real-time 3d reconstruction and 6-dof tracking with an event camera.
\newblock In {\em ECCV}, 2016.

\bibitem{Kingma14}
D.~Kingma and J.~Ba.
\newblock Adam: A method for stochastic optimization.
\newblock {\em arXiv preprint arXiv:1412.6980}, 2014.

\bibitem{Krizhevsky09}
A.~Krizhevsky and G.~Hinton.
\newblock Learning multiple layers of features from tiny images.
\newblock 2009.

\bibitem{Krizhevsky12}
A.~Krizhevsky, I.~Sutskever, and G.~Hinton.
\newblock {{ImageNet} Classification with Deep Convolutional Neural Networks}.
\newblock In {\em NIPS}, 2012.

\bibitem{Lagorce15a}
X.~Lagorce, S.-H. Ieng, X.~Clady, M.~Pfeiffer, and R.~B. Benosman.
\newblock Spatiotemporal features for asynchronous event-based data.
\newblock {\em Frontiers in neuroscience}, 2015.

\bibitem{Lagorce17}
X.~Lagorce, G.~Orchard, F.~Galluppi, B.~E. Shi, and R.~B. Benosman.
\newblock Hots: a hierarchy of event-based time-surfaces for pattern
  recognition.
\newblock {\em TPAMI}, 2017.

\bibitem{Laptev05}
I.~Laptev.
\newblock On space-time interest points.
\newblock {\em IJCV}, 2005.

\bibitem{Lecun89}
Y.~LeCun, B.~E. Boser, J.~S. Denker, D.~Henderson, R.~E. Howard, W.~E. Hubbard,
  and L.~D. Jackel.
\newblock Handwritten digit recognition with a back-propagation network.
\newblock In {\em NIPS}, 1989.

\bibitem{Lee16}
J.~H. Lee, T.~Delbruck, and M.~Pfeiffer.
\newblock Training deep spiking neural networks using backpropagation.
\newblock {\em Frontiers in neuroscience}, 2016.

\bibitem{Li16}
H.~Li, G.~Li, and L.~Shi.
\newblock Classification of spatiotemporal events based on random forest.
\newblock In {\em Advances in Brain Inspired Cognitive Systems: International
  Conference}, 2016.

\bibitem{Li17}
H.~Li, H.~Liu, X.~Ji, G.~Li, and L.~Shi.
\newblock Cifar10-dvs: An event-stream dataset for object classification.
\newblock {\em Frontiers in neuroscience}, 11:309, 2017.

\bibitem{Lichtsteiner08}
P.~Lichtsteiner, C.~Posch, and T.~Delbruck.
\newblock A 128x128 120db 15us latency asynchronous temporal contrast vision
  sensor.
\newblock {\em IEEE Journal of Solid State Circuits}, 2008.

\bibitem{Lin14}
T.-Y. Lin, M.~Maire, S.~Belongie, J.~Hays, P.~Perona, D.~Ramanan,
  P.~Doll{\'a}r, and C.~Zitnick.
\newblock {Microsoft COCO: Common Objects in Context}.
\newblock In {\em ECCV}, 2014.

\bibitem{Linares11}
B.~Linares-Barranco, T.~Serrano-Gotarredona, L.~A. Camu{\~n}as-Mesa, J.~A.
  Perez-Carrasco, C.~Zamarre{\~n}o-Ramos, and T.~Masquelier.
\newblock On spike-timing-dependent-plasticity, memristive devices, and
  building a self-learning visual cortex.
\newblock {\em Frontiers in neuroscience}, 2011.

\bibitem{Liu16}
H.~Liu, D.~P. Moeys, G.~Das, D.~Neil, S.-C. Liu, and T.~Delbr{\"u}ck.
\newblock Combined frame-and event-based detection and tracking.
\newblock In {\em Circuits and Systems, 2016 IEEE International Symposium on},
  2016.

\bibitem{Lowe99}
D.~G. Lowe.
\newblock Object recognition from local scale-invariant features.
\newblock In {\em ICCV}, 1999.

\bibitem{Marti16}
D.~Mart{\'\i}, M.~Rigotti, M.~Seok, and S.~Fusi.
\newblock Energy-efficient neuromorphic classifiers.
\newblock {\em Neural computation}, 2016.

\bibitem{Masquelier07}
T.~Masquelier and S.~J. Thorpe.
\newblock Unsupervised learning of visual features through spike timing
  dependent plasticity.
\newblock {\em PLoS computational biology}, 2007.

\bibitem{Mueggler17}
E.~Mueggler, C.~Bartolozzi, and D.~Scaramuzza.
\newblock Fast event-based corner detection.
\newblock In {\em BMVC}, 2017.

\bibitem{Mueggler17b}
E.~Mueggler, H.~Rebecq, G.~Gallego, T.~Delbruck, and D.~Scaramuzza.
\newblock The event-camera dataset and simulator: Event-based data for pose
  estimation, visual odometry, and slam.
\newblock {\em The International Journal of Robotics Research}, 2017.

\bibitem{Neil16b}
D.~Neil, M.~Pfeiffer, and S.-C. Liu.
\newblock Learning to be efficient: Algorithms for training low-latency,
  low-compute deep spiking neural networks.
\newblock In {\em Proceedings of the 31st Annual ACM Symposium on Applied
  Computing}. ACM, 2016.

\bibitem{Neil16}
D.~Neil, M.~Pfeiffer, and S.-C. Liu.
\newblock Phased lstm: Accelerating recurrent network training for long or
  event-based sequences.
\newblock In {\em NIPS}, 2016.

\bibitem{Oconnor13}
P.~O'Connor, D.~Neil, S.-C. Liu, T.~Delbruck, and M.~Pfeiffer.
\newblock {Real-time classification and sensor fusion with a spiking deep
  belief network.}
\newblock {\em Frontiers in neuroscience}, 2013.

\bibitem{Orchard15}
G.~Orchard, A.~Jayawant, G.~K. Cohen, and N.~Thakor.
\newblock Converting static image datasets to spiking neuromorphic datasets
  using saccades.
\newblock {\em Frontiers in Neuroscience}, 2015.

\bibitem{Orchard14}
G.~Orchard, D.~Matolin, X.~Lagorce, R.~Benosman, and C.~Posch.
\newblock Accelerated frame-free time-encoded multi-step imaging.
\newblock In {\em Circuits and Systems, 2014 IEEE International Symposium on},
  2014.

\bibitem{Orchard15c}
G.~Orchard, C.~Meyer, R.~Etienne-Cummings, C.~Posch, N.~Thakor, and
  R.~Benosman.
\newblock Hfirst: A temporal approach to object recognition.
\newblock {\em TPAMI}, 2015.

\bibitem{Peng17}
X.~Peng, B.~Zhao, R.~Yan, H.~Tang, and Z.~Yi.
\newblock Bag of events: An efficient probability-based feature extraction
  method for aer image sensors.
\newblock {\em IEEE transactions on neural networks and learning systems},
  2017.

\bibitem{Posch11}
C.~Posch, D.~Matolin, and R.~Wohlgenannt.
\newblock {A QVGA 143 dB Dynamic Range Frame-Free PWM Image Sensor With
  Lossless Pixel-Level Video Compression and Time-Domain CDS}.
\newblock {\em Solid-State Circuits, IEEE Journal of}, 2011.

\bibitem{Posch14}
C.~Posch, T.~Serrano-Gotarredona, B.~Linares-Barranco, and T.~Delbruck.
\newblock Retinomorphic event-based vision sensors: {Bioinspired} cameras with
  spiking output.
\newblock {\em Proceedings of the IEEE}, 2014.

\bibitem{Rebecq17}
H.~Rebecq, T.~Horstschaefer, G.~Gallego, and D.~Scaramuzza.
\newblock Evo: A geometric approach to event-based 6-dof parallel tracking and
  mapping in real time.
\newblock {\em IEEE Robotics and Automation Letters}, 2017.

\bibitem{Redmon16}
J.~Redmon and A.~Farhadi.
\newblock {YOLO9000:} better, faster, stronger.
\newblock {\em CoRR}, 2016.

\bibitem{ren15}
S.~Ren, K.~He, R.~Girshick, and J.~Sun.
\newblock Faster {R-CNN}: Towards real-time object detection with region
  proposal networks.
\newblock In {\em NIPS}, 2015.

\bibitem{Rublee11}
E.~Rublee, V.~Rabaud, K.~Konolige, and G.~Bradski.
\newblock Orb: An efficient alternative to sift or surf.
\newblock In {\em ICCV}, 2011.

\bibitem{Rueckauer16}
B.~Rueckauer, I.-A. Lungu, Y.~Hu, and M.~Pfeiffer.
\newblock Theory and tools for the conversion of analog to spiking
  convolutional neural networks.
\newblock {\em arXiv preprint arXiv:1612.04052}, 2016.

\bibitem{Russell10}
A.~Russell, G.~Orchard, Y.~Dong, {\c{S}}.~Mihalas, E.~Niebur, J.~Tapson, and
  R.~Etienne-Cummings.
\newblock Optimization methods for spiking neurons and networks.
\newblock {\em IEEE transactions on neural networks}, 2010.

\bibitem{Scherer10}
M.~Scherer, M.~Walter, and T.~Schreck.
\newblock Histograms of oriented gradients for 3d object retrieval.
\newblock In {\em WSCG}, 2010.

\bibitem{Seifozzakerini16}
S.~Seifozzakerini, W.-Y. Yau, B.~Zhao, and K.~Mao.
\newblock Event-based hough transform in a spiking neural network for multiple
  line detection and tracking using a dynamic vision sensor.
\newblock In {\em BMVC}, 2016.

\bibitem{Serrano06}
R.~Serrano-Gotarredona, T.~Serrano-Gotarredona, A.~Acosta-Jimenez, and
  B.~Linares-Barranco.
\newblock A neuromorphic cortical-layer microchip for spike-based event
  processing vision systems.
\newblock {\em IEEE Transactions on Circuits and Systems I: Regular Papers},
  2006.

\bibitem{Serrano13}
T.~Serrano-Gotarredona and B.~Linares-Barranco.
\newblock A 128 x 128 1.5\% contrast sensitivity 0.9\% fpn 3 $\mu$s latency 4
  mw asynchronous frame-free dynamic vision sensor using transimpedance
  preamplifiers.
\newblock {\em Solid-State Circuits, IEEE Journal of}, 2013.

\bibitem{Serrano15}
T.~Serrano-Gotarredona and B.~Linares-Barranco.
\newblock Poker-dvs and mnist-dvs. their history, how they were made, and other
  details.
\newblock {\em Frontiers in neuroscience}, 2015.

\bibitem{Sheik13}
S.~Sheik, M.~Pfeiffer, F.~Stefanini, and G.~Indiveri.
\newblock Spatio-temporal spike pattern classification in neuromorphic systems.
\newblock In {\em Biomimetic and Biohybrid Systems}. 2013.

\bibitem{Shelhamer17}
E.~Shelhamer, J.~Long, and T.~Darrell.
\newblock Fully convolutional networks for semantic segmentation.
\newblock {\em TPAMI}, 2017.

\bibitem{Sivic09}
J.~Sivic and A.~Zisserman.
\newblock Efficient visual search of videos cast as text retrieval.
\newblock {\em TPAMI}, 2009.

\bibitem{Srivastava14}
N.~Srivastava, G.~E. Hinton, A.~Krizhevsky, I.~Sutskever, and R.~Salakhutdinov.
\newblock Dropout: a simple way to prevent neural networks from overfitting.
\newblock {\em Journal of machine learning research}, 2014.

\bibitem{Stromatias17}
E.~Stromatias, M.~Soto, T.~Serrano-Gotarredona, and B.~Linares-Barranco.
\newblock An event-driven classifier for spiking neural networks fed with
  synthetic or dynamic vision sensor data.
\newblock {\em Frontiers in neuroscience}, 2017.

\bibitem{Tan15}
C.~Tan, S.~Lallee, and G.~Orchard.
\newblock Benchmarking neuromorphic vision: lessons learnt from computer
  vision.
\newblock {\em Frontiers in neuroscience}, 2015.

\bibitem{Vasco16}
V.~Vasco, A.~Glover, and C.~Bartolozzi.
\newblock Fast event-based harris corner detection exploiting the advantages of
  event-driven cameras.
\newblock In {\em Intelligent Robots and Systems (IROS), 2016 IEEE/RSJ
  International Conference on}, 2016.

\bibitem{Viola04}
P.~Viola and M.~J. Jones.
\newblock Robust real-time face detection.
\newblock {\em IJCV}, 2004.

\bibitem{Wang09}
H.~Wang, M.~M. Ullah, A.~Klaser, I.~Laptev, and C.~Schmid.
\newblock Evaluation of local spatio-temporal features for action recognition.
\newblock In {\em BMVC}, 2009.

\bibitem{Witkin84}
A.~Witkin.
\newblock Scale-space filtering: A new approach to multi-scale description.
\newblock In {\em Acoustics, Speech, and Signal Processing, IEEE International
  Conference on.}, 1984.

\bibitem{Xie17}
S.~Xie and Z.~Tu.
\newblock Holistically-nested edge detection.
\newblock {\em IJCV}, 2017.

\bibitem{Zhao15}
B.~Zhao, R.~Ding, S.~Chen, B.~Linares-Barranco, and H.~Tang.
\newblock Feedforward categorization on aer motion events using cortex-like
  features in a spiking neural network.
\newblock {\em IEEE transactions on neural networks and learning systems},
  2015.

\end{thebibliography}
}

\end{document}